\documentclass[preprint,12pt]{elsarticle}

\usepackage{graphicx}
\usepackage{multirow}
\usepackage{amsmath,amssymb,amsfonts}
\usepackage{amsthm}
\usepackage{mathrsfs}
\usepackage[title]{appendix}
\usepackage{xcolor}
\usepackage{textcomp}
\usepackage{manyfoot}
\usepackage{booktabs}
\usepackage{algorithm}
\usepackage{algorithmicx}
\usepackage{algpseudocode}
\usepackage{listings}
\biboptions{numbers,sort&compress}

\journal{arXiv}
\date{}

\begin{document}

\begin{frontmatter}

\title{Attention Guidance Mechanism for Handwritten Mathematical Expression Recognition}

\author[1]{Yutian Liu\fnref{coauthor}}\ead{liuyutian@tju.edu.cn}
\author[1]{Wenjun Ke\fnref{coauthor}}\ead{wenjunke@tju.edu.cn}
\author[1]{Jianguo Wei\corref{cor}}\ead{jianguo@tju.edu.cn}
\affiliation[1]{organization={College of Intelligence and Computing, Tianjin University},
                postcode={300000},
                city={Tianjin},
                country={China}}
\fntext[coauthor]{Both authors contributed equally}
\cortext[cor]{Corresponding author}

\begin{abstract}
Handwritten mathematical expression recognition (HMER) is challenging in image-to-text tasks due to the complex layouts of mathematical expressions and suffers from problems including over-parsing and under-parsing. To solve these, previous HMER methods improve the attention mechanism by utilizing historical alignment information. However, this approach has limitations in addressing under-parsing since it cannot correct the erroneous attention on image areas that should be parsed at subsequent decoding steps. This faulty attention causes the attention module to incorporate future context into the current decoding step, thereby confusing the alignment process. To address this issue, we propose an attention guidance mechanism to explicitly suppress attention weights in irrelevant areas and enhance the appropriate ones, thereby inhibiting access to information outside the intended context. Depending on the type of attention guidance, we devise two complementary approaches to refine attention weights: self-guidance that coordinates attention of multiple heads and neighbor-guidance that integrates attention from adjacent time steps. Experiments show that our method outperforms existing state-of-the-art methods, achieving expression recognition rates of 60.75\% / 61.81\% / 63.30\% on the CROHME 2014/ 2016/ 2019 datasets.
\end{abstract}


\end{frontmatter}

\section{Introduction}\label{sec1}

Handwritten mathematical expression recognition (HMER) aims to translate mathematical expression images into corresponding \LaTeX\ sequences. HMER is a fundamental OCR task in educational scenarios, which has led to many real-world applications, including automatic scoring and online education. However, HMER is challenging because mathematical expressions are laid out in two dimensions, different from natural language texts including English, Chinese, etc. An HMER algorithm is expected to analyze the structure of an expression while recognizing the symbols \cite{IJDAR2012a}. Over the past few years, HMER methods have benefited from deep neural networks and have made significant progress \cite{WAP, DWAP, TD, BTTR, CoMER, TDv2, ABM, CAN, GCN, EMNLP23, TMM23}. Most prevalent methods treat HMER as a sequence-to-sequence task and follow the attention-based encoder-decoder framework. In this framework, the encoder first extracts the visual features of the input image as a sequence, and then the decoder generates the target sequence by attending to the encoded feature sequence \cite{WAP, TD, BTTR} at each decoding step. The attention mechanism attends to image regions at each decoding step based on the hidden state of the last step.

\begin{figure}
    \centering
    \includegraphics[width=0.60\linewidth]{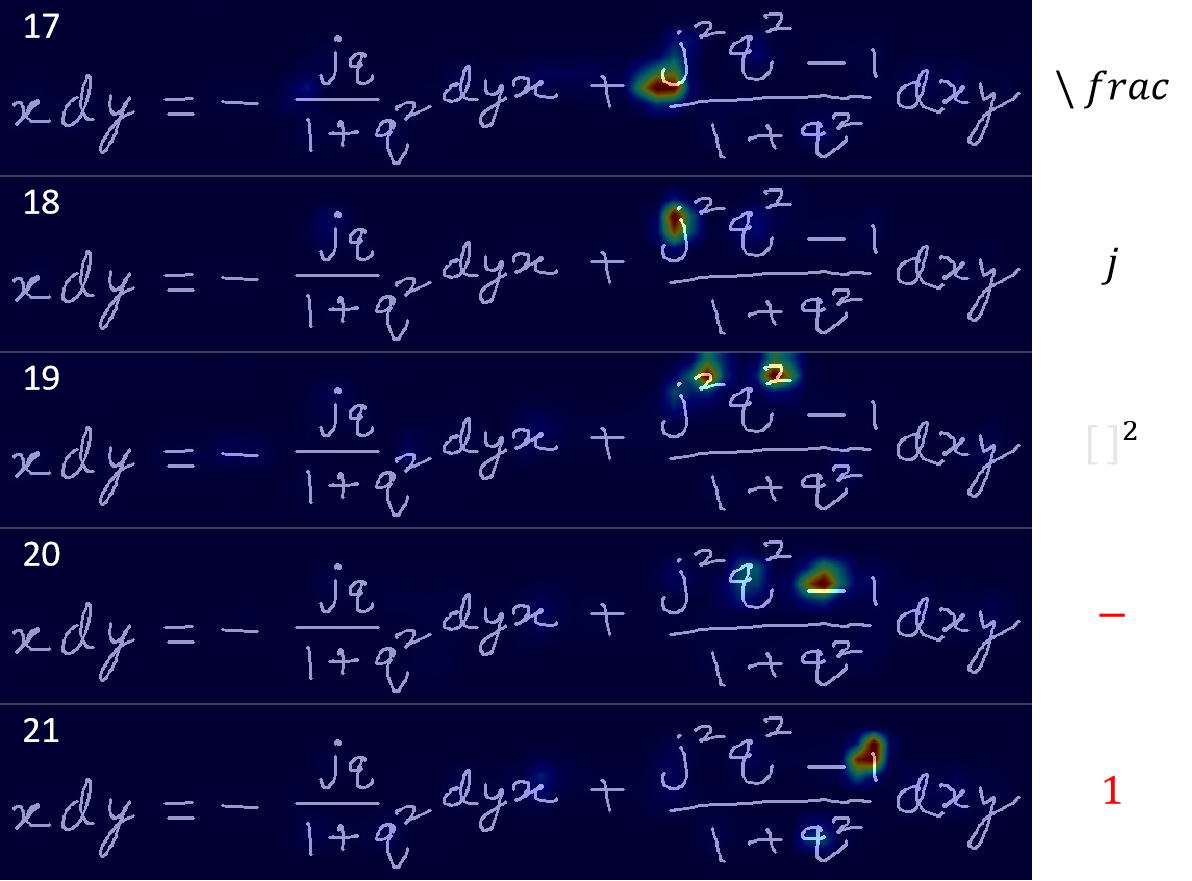}
    \caption{Illustration of the context leakage phenomenon observed in a state-of-the-art coverage-based system \cite{CoMER}. The rows are a series of attention maps attending to visible symbols. We skip attention maps of structural symbols (``\{'', ``\}'', ``\^{}'' and ``\_'') for simplicity. The number in the upper left corner of each row denotes the $i$-th visible symbol in the generated sequence.}
    \label{fig:1}
\end{figure}

Beyond the hidden state, many HMER methods \cite{WAP, CoMER, DWAP, ABM, CAN} exploit historical attention weights to prompt the attention module to focus on unattended image areas, called the coverage mechanism. The coverage mechanism is proposed to solve the over-parsing and under-parsing problems. Over-parsing occurs when a particular symbol region is parsed multiple times unnecessarily, whereas under-parsing happens when a symbol region is overlooked and not parsed at all \cite{WAP}. The coverage mechanism successfully alleviates over-parsing but has limitations in addressing under-parsing. Fig. \ref{fig:1} illustrates a series of attended image regions from a state-of-the-art coverage-based system \cite{CoMER}. As we can see, the superscripts of ``$j^{2}$'' and ``$q^{2}$'' are activated at the same time. As a result, features of another ``2'' that should be decoded in the future are incorporated in the current context, causing ``$q^{2}$'' to be omitted in the decoding process.

We refer to the above observation as the \textit{context leakage} phenomenon: if the attention module attends to image areas that should be parsed at subsequent decoding steps, context from the future will be incorporated into the attention result, causing symbols between the target region and the incorrectly attended areas to tend to be omitted. The coverage mechanism cannot solve this phenomenon since it is only aware of past alignment information. Different from the multi-activation phenomenon in \cite{ABINetPP, LISTER}, the context leakage phenomenon emphasizes the information leakage caused by the early activation of image areas that should be parsed after the target region.

To address this issue, we propose an \textit{attention guidance mechanism} which aims to explicitly suppress attention weights in irrelevant areas and enhance ones in the target region. This mechanism comprises two complementary approaches: self-guidance and neighbor-guidance.

\textit{Self-guidance} aims to refine correlation maps by eliminating inconsistencies of different attention heads. In multi-head attention, each head learns its own set of linear projection matrices for queries, keys, and values. This independence allows the model to capture a richer variety of information from different perspectives. Although different heads attend to image areas in different subspaces, the attended symbol in the input image should be consistent across multiple heads. However, due to symbols with similar visual features and context, certain heads may attend to different symbols. The incorrectly attended symbols are the cause of the context leakage phenomenon. Therefore, inconsistent attention should be eliminated. We achieve this through a guidance map indicating inconsistencies among attention heads and a linear projection to adjust the guidance weights.

\begin{figure}
    \centering
    \includegraphics[width=1.0\linewidth]{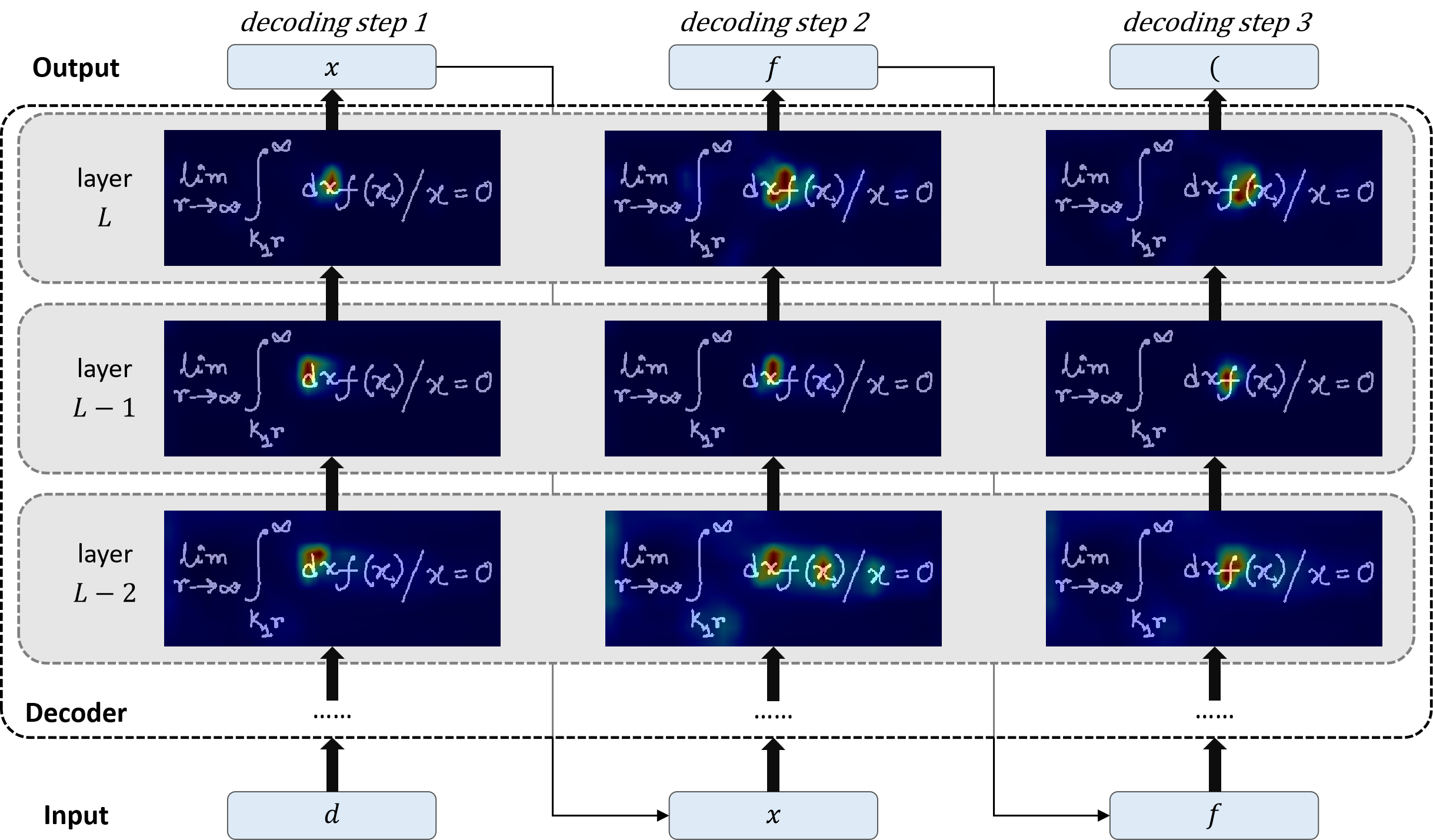}
    \caption{Alignment process of a Transformer decoder with $L$ layers. As we can see, the output layer attends to the image region of the symbol to be generated, while the middle layers focus on its previously decoded neighbor. Note that the three ``$x$'' are attended simultaneously in layer $L-2$ at decoding step 2, which causes the context leakage phenomenon.}
    \label{fig:2}
\end{figure}

While self-guidance is performed in each single decoding step, attention from adjacent time steps can also be involved in the attention guidance mechanism. For HMER methods with a stacked decoder \cite{StackedDecoder, BTTR, CoMER, GCN}, the query of the attention module in each decoder layer is determined by the output of its previous layer. Fig. \ref{fig:2} illustrates how a stacked decoder generates the final attention regions. As we can see, only the output layer focuses on the symbol to be generated. The middle layers first calculate attention weights based on the previous symbol instead of directly aligning with the current one. In other words, each symbol is aligned via its previously decoded neighbor. Different HMER methods \cite{BTTR, CoMER} with different decoder layers ($L \geq 2$) follow this process consistently. Moreover, a similar phenomenon has also been found in image captioning systems \cite{AoA, ProphetAttn}.

Based on the above observation, we propose another attention guidance approach, \textit{neighbor-guidance}. Neighbor-guidance aims to leverage the final attention weights of the previous decoding step to enhance appropriate attention for middle layers at the current step. For example, in Fig. \ref{fig:2}, attention weights for generating ``$x$'' in layer $L$ can be reused to refine the attention weights of layer $L-2$ at decoding step 2, where the three ``$x$'' are attended simultaneously and the context leakage phenomenon occurs. With the help of neighbor-guidance, appropriate attention weights of middle layers are enhanced, thus improving the attention quality of both the current and final layers. Neighbor-guidance is implemented similarly to self-guidance but changes the attention guidance from self-generated attention to the attention of the last decoding step and can be readily employed without training. Moreover, different from self-guidance, neighbor-guidance can be applied only to middle layers and facilitates information propagation between adjacent time steps.

Overall, the contributions of this work are as follows:

\begin{itemize}
    \item We propose the phenomenon of context leakage, which cannot be solved by the widely used coverage mechanism. To solve this issue, we propose a general attention guidance mechanism to explicitly suppress attention weights in irrelevant areas and enhance ones in the target region.
    \item We devise self-guidance to refine the raw attention by seeking consensus among multiple attention heads, based on the intuition that the attended symbol in the input image should be consistent for all attention heads at each decoding step.
    \item We devise neighbor-guidance to refine the raw attention by reusing attention weights from the last decoding step, based on the observation that the alignment of the current symbol relies on that of its previously decoded neighbor.
    \item Experiments on standard benchmarks show that our method outperforms state-of-the-art methods, achieving expression recognition rates (ExpRate)s of 60.75\% / 61.81\% / 63.30\% on the CROHME 2014 \cite{CROHME2014}/ 2016 \cite{CROHME2016}/ 2019 \cite{CROHME2019} datasets.
\end{itemize}

\section{Related work}\label{sec2}

The section below describes related works of HMER. Traditional HMER methods involve explicit symbol recognition and layout analysis, while deep learning-based methods generate the target sequence through an end-to-end encoder-decoder network.

\subsection{Traditional HMER methods}\label{sec2-1}

Traditional HMER methods are considered to be the combination of symbol recognition and layout analysis by existing literature reviews \cite{IJDAR2012a, IJDAR2000}. According to the implementation of these two parts, traditional methods can be divided into sequential and global methods.

\textit{Sequential methods} \cite{PAMI2002, PRL2014a} perform symbol segmentation, symbol recognition, and two-dimensional layout analysis sequentially, where each step depends on the result of the previous step. The downside of such methods is the lack of utilization of global information and error accumulation. In contrast, \textit{global methods} \cite{PRL2014b, PR2016} recognize symbols and analyze two-dimensional layouts simultaneously. In this way, symbol recognition and layout analysis can be optimized through the global information of the input expression, and symbol segmentation is achieved as a by-product of the optimization. Unfortunately, the computational cost of these methods grows exponentially with the number of symbols.

On the other hand, both sequential and global methods usually require a predefined grammar, such as graph grammar \cite{IJDAR2020}, relational grammar \cite{IJDAR2012b}, and context-free grammar \cite{PRL2014a}. These grammars are constructed based on extensive prior knowledge. Overall, traditional HMER methods have limitations including explicit segmentation, prior knowledge to define grammars, and the exponential computational complexity of the parsing algorithms.

\subsection{Deep learning-based HMER methods}\label{sec2-2}

Deep learning-based HMER methods avoid the limitations of traditional approaches. Next, we comment on a few deep learning-based HMER methods from three aspects: model architectures, attention mechanisms, and training strategies.

\textit{Model architectures}. Over the past few years, various image-to-text tasks have benefited from the encoder-decoder architecture, such as image captioning \cite{ShowAttendTell}, scene text recognition \cite{FAN-STR}, and handwritten text recognition \cite{ONN-HTR}. Inspired by these advances, Deng et al. \cite{Coarse2Fine} first introduced the encoder-decoder framework into the offline mathematical expression recognition task. Afterward, a series of important HMER methods have been proposed \cite{WAP, DWAP}, achieving excellent results on CROHME benchmarks. They are equipped with a convolution neural network (CNN) encoder and a gated recurrent unit (GRU) \cite{GRU} decoder, where \cite{DWAP} uses a multi-scale DenseNet \cite{DenseNet} encoder to replace the VGG \cite{VGG} architecture in \cite{WAP}.

As Transformer \cite{Transformer} has gradually replaced RNN as the preferred sequence model in various tasks, recent methods have attempted to apply Transformer to the HMER task. In CROHME 2019, Uniz. Linz \cite{CROHME2019} employed a Faster R-CNN \cite{FasterRCNN} detector and a Transformer decoder, where the two parts were trained separately. Subsequently, Zhao et al. \cite{BTTR} proposed a model named BTTR, which first introduced the end-to-end Transformer decoder to solve the HMER task. The encoder of BTTR is the same as \cite{DWAP}, except that an additional $1 \times 1$ convolution layer is added to adjust the image feature dimension, and a 2-D positional encoding is added to represent image positional features. Inspired by the design of Transformer, \cite{StackedDecoder} proposed to use multi-head attention and a stacked decoder to improve the performance of RNN-based models.

In addition to the methods mentioned above that directly generate target sequences, many methods produce results with complex structures. For example, Zhang et al. \cite{TD} proposed a tree-structured decoder (TD), where a parent node, a child node, and their relationship are generated simultaneously at each step. These nodes are consequently used to construct the final tree structure. Furthermore, TDv2 \cite{TDv2} was proposed to utilize the tree structure labels fully. Specifically, its decoder consists of a symbol classification module and a relationship prediction module. These methods also consist of a CNN-based encoder and an RNN-based decoder, even though they produce results with different structures.

\textit{Attention mechanisms}. As a core component of the encoder-decoder framework, attention mechanisms have attracted much research interest in the HMER community. Zhang et al. \cite{WAP} first introduced the coverage mechanism to solve the under-parsing and over-parsing problems in HMER. Under-parsing means that some parts of the image are omitted from the decoding, while over-parsing implies that certain areas are unnecessarily parsed multiple times in the decoding process. The coverage mechanism uses historical alignment information to prompt the attention module to focus on unattended regions. With the excellent performance of the coverage mechanism, it has been widely used in subsequent RNN-based HMER methods \cite{ABM, CAN, DWAP, StackedDecoder, ScaleAug, PAL}. However, the coverage mechanism cannot be applied directly to the Transformer decoder due to its parallel computing. CoMER \cite{CoMER} made it possible to compute coverage attention in a Transformer decoder without harming its parallel decoding nature and achieved the leading ExpRate on three CROHME benchmarks. Moreover, Zhang et al. \cite{DWAP} proposed a multi-scale attention module that deals with recognizing symbols in different scales. Similarly, Bian et al. \cite{ABM} proposed an attention aggregation module to integrate multi-scale coverage attention.

\textit{Training strategies}. Various weakly supervised information have been introduced to improve the performance of HMER methods. For example, Truong et al. \cite{WS-WAP} introduced information about the absence or presence of symbols to the encoder. Li et al. \cite{CAN} proposed a weakly supervised counting module and jointly optimized HMER and symbol counting tasks. Regarding bidirectional training, Bian et al. \cite{ABM} proposed ABM that consists of two parallel inverse decoders, which were enhanced via mutual distillation. Besides, BTTR \cite{BTTR} applied a bidirectional training strategy to the Transformer decoder for bidirectional language modeling. For data augmentation, Li et al. \cite{ScaleAug} proposed a scale augmentation method for HMER, which scales input images randomly while maintaining their aspect ratio. PAL \cite{PAL} was proposed to use printed mathematical expressions as extra data to help train the model via paired adversarial learning, which maps handwritten features into printed ones.

\section{Proposed method}\label{sec3}

In this section, we first describe the overall architecture of our HMER method based on \cite{CoMER} and then introduce the proposed attention guidance mechanism, which is an extension of the cross-attention module.

\subsection{Overall architecture}\label{sec3-1}

\begin{figure}[!t]
    \centering
    \includegraphics[width=0.75\linewidth]{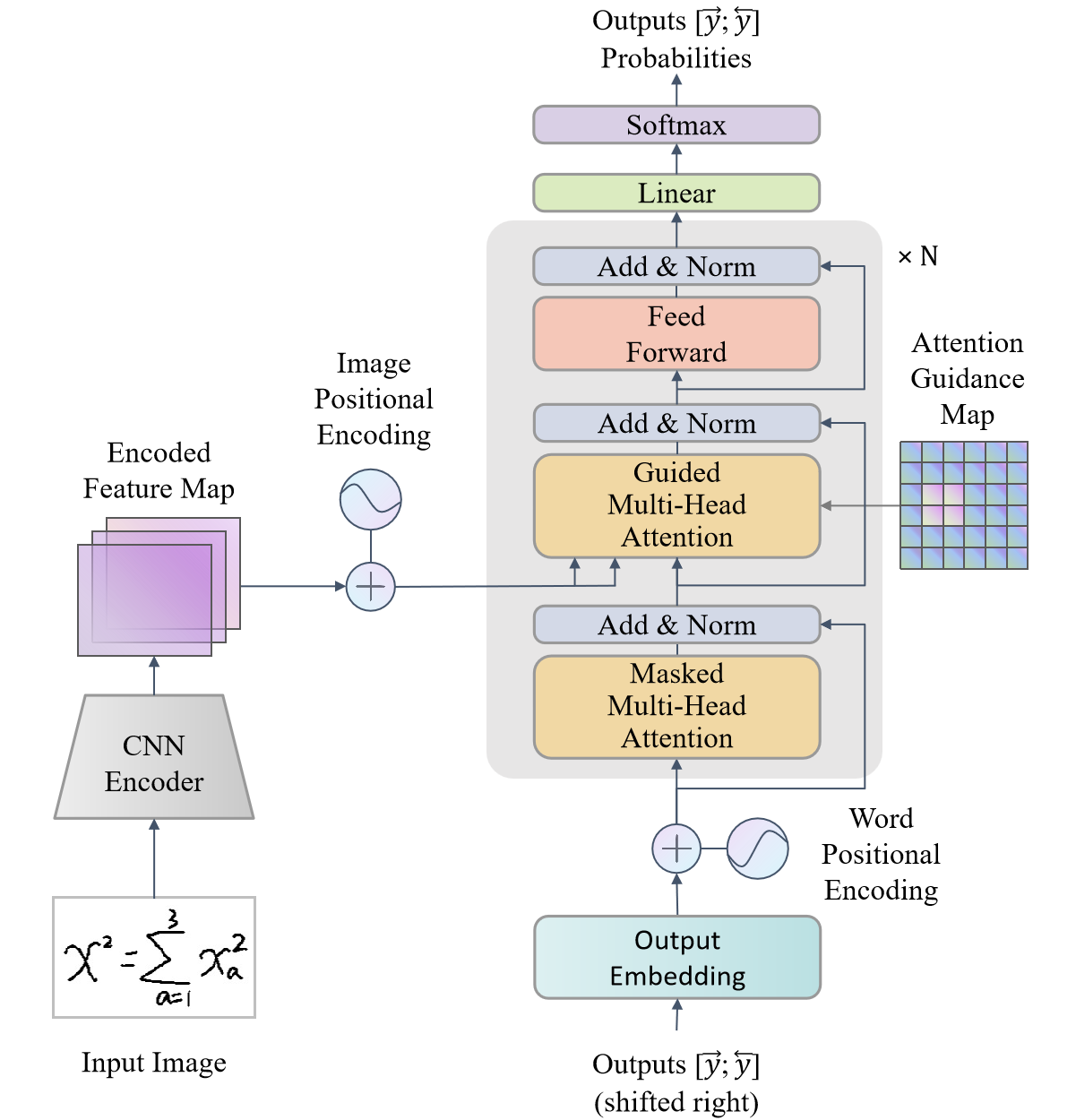}
    \caption{Handwritten mathematical expression recognition with the proposed attention guidance mechanism. The model is based on CoMER \cite{CoMER}, which consists of a CNN encoder and a bidirectionally trained Transformer decoder. The bidirectional decoder receives input and produces output in two directions simultaneously. Attention guidance is applied to the cross-attention modules of the decoder, where attention weights are refined via the guidance map.}
    \label{fig:3}
\end{figure}

An overview of the HMER model is presented in Fig. \ref{fig:3}. The model comprises a CNN encoder and a Transformer decoder with the proposed attention guidance mechanism.

\subsubsection{CNN encoder}

For the encoder part, we use DenseNet \cite{DenseNet} to extract visual features from input images, following the same setting with \cite{BTTR} and \cite{CoMER}. The core principle of DenseNet is to ensure that each layer receives feature maps from all preceding layers as inputs. This is achieved through dense blocks, where the output of each layer within a block is concatenated with the inputs of all subsequent layers. Specifically, in the dense block $b$, the output of the $l^{th}$ layer can be computed as: 
\begin{equation}
    \mathbf{X}_l=H_l([\mathbf{X}_0;\mathbf{X}_1;\ldots;\mathbf{X}_{l-1}])\in\mathbb{R}^{h_b\times w_b\times d_b}
\end{equation} 
where $\mathbf{X}_0,\mathbf{X}_1,\ldots,\mathbf{X}_{l}\in\mathbb{R}^{h_b\times w_b\times d_b}$ denote the output feature maps from the $0^{th}$ to $l^{th}$ layers, ``;'' denotes the concatenation operation, $d_{b}$ denotes the feature dimension of the dense block $b$, and the convolution function $H_{l}(\cdot)$ consists of a $3\times3$ convolution layer, a ReLU \cite{ReLU} activation function, and a batch normalization \cite{BatchNorm}. This mechanism enhances the flow of information and gradients throughout the network.

After the encoding stage, the encoded feature map is then projected to the model dimension $d$ of the decoder. Furthermore, the feature map adds an image positional encoding before interacting with the decoder, which is the same as \cite{CoMER, DETR}. This spatial awareness is crucial for the HMER task since the position of symbols within the image is essential for the decoding process. To compute the positional encoding, the position coordinates are first normalized as relative positions: 
\begin{equation}
    \bar{x}=\frac x{w_o},\quad\bar{y}=\frac y{h_o}
\end{equation} 
where $(x,y)$ is a 2D coordinates tuple, and $(h_0,w_0)$ denotes the shape of the feature map $\mathbf{X}\in\mathbb{R}^{h_o\times w_o\times d}$. The image positional encoding is then represented as the concatenation of 1D word positional encoding \cite{Transformer}. 
\begin{equation}
    \mathbf{p}_{x,y,d}^{2D}=[\mathbf{p}_{\bar{x},d/2}^{1D};\mathbf{p}_{\bar{y},d/2}^{1D}]\in\mathbb{R}^{d}
\end{equation}
\begin{equation}\begin{aligned}
    \mathbf{p}_{p,d}^{1D}[2i]&=\sin(p/10000^{2i/d})\\\mathbf{p}_{p,d}^{1D}[2i+1]&=\cos(p/10000^{2i/d})
\end{aligned}\end{equation}

\subsubsection{Transformer decoder}

Transformer decoder has been widely used in sequence generation tasks \cite{VisualGPT, TROCR, CoMER} due to its effectiveness in capturing complex dependencies within the input sequence and between the input and output sequences. The Transformer decoder architecture is characterized by a stack of identical layers comprising a self-attention module, a cross-attention module, and a feed-forward layer, as shown in Fig \ref{fig:3}.

As the basis of self-attention and cross-attention, multi-head attention is the most critical component of the transformer decoder. Given a set of queries $\mathbf{Q}$, keys $\mathbf{K}$, and values $\mathbf{V}$, the attention mechanism computes the output as a weighted sum of the values, where the weight assigned to each value is determined by a compatibility function of the query with the corresponding key. For each head $\mathrm{Head}_i$, this process is formalized by the scaled dot-product attention formula: 
\begin{equation}
    \mathbf{Q}_i,\mathbf{K}_i,\mathbf{V}_i=\mathbf{Q}\mathbf{W}_i^Q,\mathbf{K}\mathbf{W}_i^K,\mathbf{V}\mathbf{W}_i^V
\end{equation}
\begin{equation}\begin{aligned}\label{DotProduct}
    \mathbf{E}_i=\frac{\mathbf{Q}_i\mathbf{K}_i^\intercal}{\sqrt{d_k}}\in\mathbb{R}^{T\times L}
\end{aligned}\end{equation}
\begin{equation}\label{Softmax}
    \mathbf{A}_i=\operatorname{softmax}(\mathbf{E}_i)\in\mathbb{R}^{T\times L}
\end{equation}
\begin{equation}
    \mathrm{Head}_i=\mathbf{A}_i\mathbf{V}_i\in\mathbb{R}^{T\times d_v}
\end{equation}
where $\mathbf{W}_i^Q\in\mathbb{R}^{d_\text{model}\times d_k},\mathbf{W}_i^K\in\mathbb{R}^{d_\text{model}\times d_k},\mathbf{W}_i^V\in\mathbb{R}^{d_\text{model}\times d_v}$ are trainable weights, $\mathbf{E}_i,\mathbf{A}_i$ are the correlation map and the attention map, $d_k,d_v$ and $T,L$ denote the dimensions and lengths of keys and values, respectively. All $h$ heads are then concatenated to form the final output:
\begin{equation}
    \mathrm{MultiHead}(\mathbf{Q},\mathbf{K},\mathbf{V})=\left[\mathrm{Head}_1;\ldots;\mathrm{Head}_h\right]\mathbf{W}^O
\end{equation}
where $\mathbf{W}^O\in\mathbb{R}^{hd_v\times d}$ is a trainable projection parameter matrix, and $d$ denotes the model dimension of the decoder.

In addition to the original cross-attention block, an attention refinement module (ARM) \cite{CoMER} is adopted to solve the over-parsing problem, as is the convention of previous HMER methods \cite{ABM, CAN, DWAP, StackedDecoder, ScaleAug, PAL}. ARM uses the accumulation of past attention maps to refine attention weights. Given the correlation map $\mathbf{E}_t$ and the attention map $\mathbf{A}_t$ at each decoding step $t$, ARM refines the correlation map using a convolution function $\phi(\cdot)$:
\begin{equation}\label{ARM}
    \mathrm{ARM}(\mathbf{E}_t,\mathbf{C}_t)=\mathbf{E}_t-\phi(\mathbf{C}_t)
\end{equation}
\begin{equation}
    \mathbf{C}_t=\sum_{k=1}^{t-1}\mathbf{A}_k\in\mathbb{R}^{L\times h}
\end{equation}
where $L=h_0\times w_0$ denotes the length of the flattened feature sequence, $h$ denotes the number of attention heads, and $\phi(\cdot)$ is implemented by: a $5\times 5$ convolution layer, a linear projection, a ReLU \cite{ReLU} activation function, and a batch normalization \cite{BatchNorm}. With the help of the coverage mechanism, the model tends to focus on unattended areas at each decoding step, avoiding unnecessarily paying attention to parsed areas.

This paper follows the above setting of multi-head attention and uses the attention guidance mechanism in Sec. \ref{sec3-2} to further enhance the correlation map $\mathbf{E}$ in Eq. \ref{DotProduct}.

\subsection{Attention guidance mechanism}\label{sec3-2}

\begin{figure}
    \centering
    \includegraphics[width=0.75\linewidth]{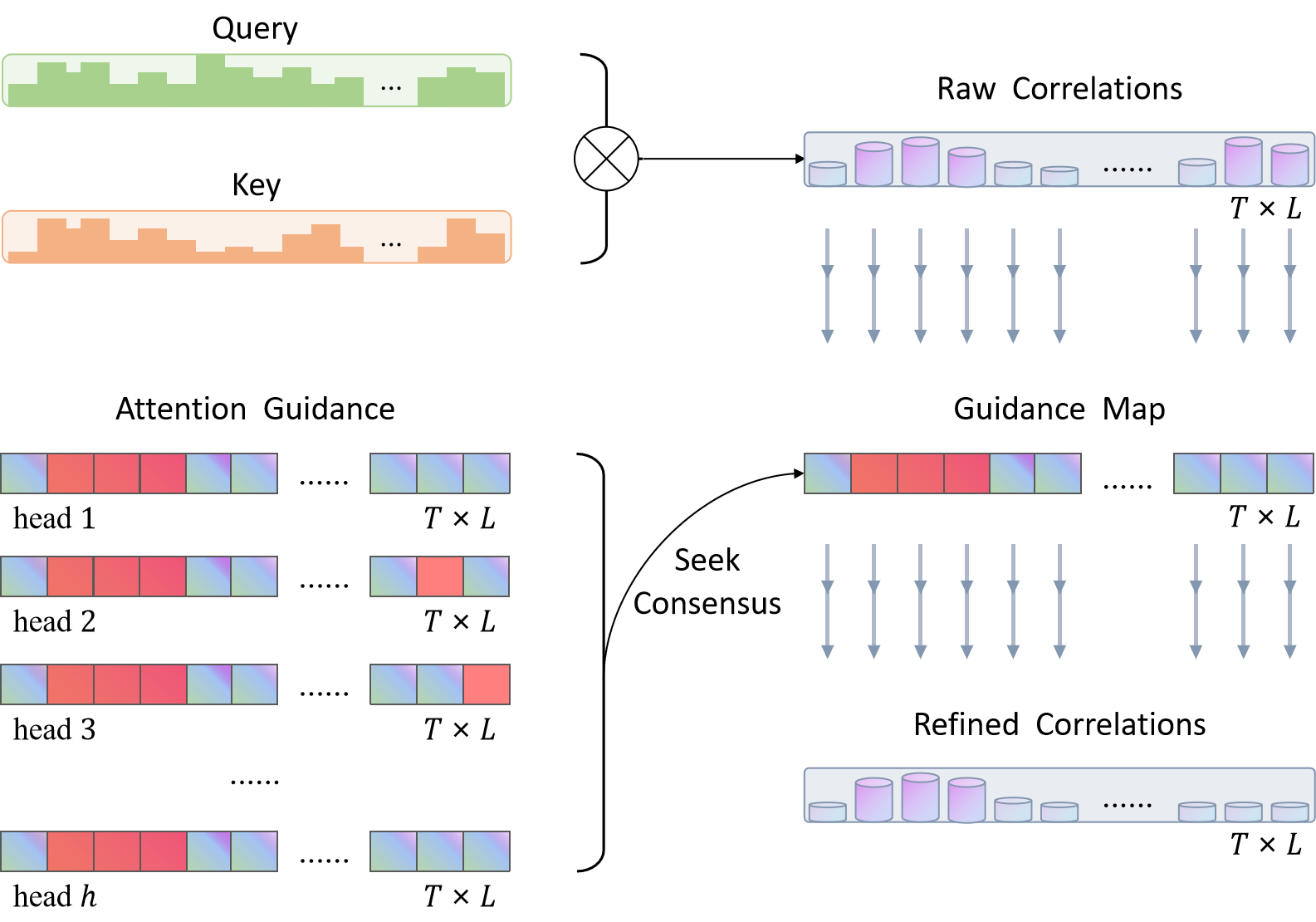}
    \caption{Attention guidance mechanism. $T$ and $L$ denote the length of the query and the key, respectively. \textit{Attention guidance} is a set of attention maps that can be derived from multiple sources. Raw correlations are refined based on the \textit{guidance map} obtained by seeking consensus from the attention guidance.}
    \label{fig:4}
\end{figure}

The proposed attention guidance mechanism aims to refine the raw correlations with the help of attention maps. It explicitly suppresses erroneous correlations, thereby avoiding misleading features of the incorrectly attended areas. Specifically, as shown in Fig. \ref{fig:4}, raw correlations $\mathbf{E}\in\mathbb{R}^{T\times L}$ calculated by Eq. \ref{DotProduct} are refined through the guidance map $\mathbf{G}\in\mathbb{R}^{T\times L}$ and a guidance function $f_{\mathrm{guide}}(\mathbf{E},\mathbf{G})$:
\begin{equation}\begin{aligned}\label{GuideFunc}
    \mathbf{\hat{E}}&=\mathbf{E}+f_{\mathrm{guide}}(\mathbf{E},\mathbf{G})\\&=\mathbf{E}+f_\mathrm{proj}(\mathbf{E}\odot\mathbf{G})
\end{aligned}\end{equation}
where $\mathbf{\hat{E}}$ denotes the refined correlations, $f_\mathrm{proj}(\cdot)$ represents a linear projection, and $\odot$ denotes the element-wise multiplication. Given a set of attention maps $\mathbf{A}^{G}\in\mathbb{R}^{T\times L\times h}$ as the guidance, $\mathbf{G}$ can be obtained by averaging attention heads or the convolution function $\phi(\cdot)$ in Eq. \ref{ARM}:
\begin{equation}\label{ConvSoftmax}
    \mathbf{G}=\mathrm{softmax}(\phi(\mathbf{A}^{G}))
\end{equation}

Note that $\mathbf{C}_t$ in Eq. \ref{ARM} can also be seen as a kind of attention guidance, as it is used to suppress activations on the past attended areas. However, it is insufficient to solve the context leakage phenomenon since it cannot deal with erroneous activations on unattended image regions. The remainder of this section introduces the two novel attention guidance approaches following Eq. \ref{GuideFunc}.

\subsubsection{Self-guidance}

\begin{figure}[!t]
    \centering
    \includegraphics[width=1.0\linewidth]{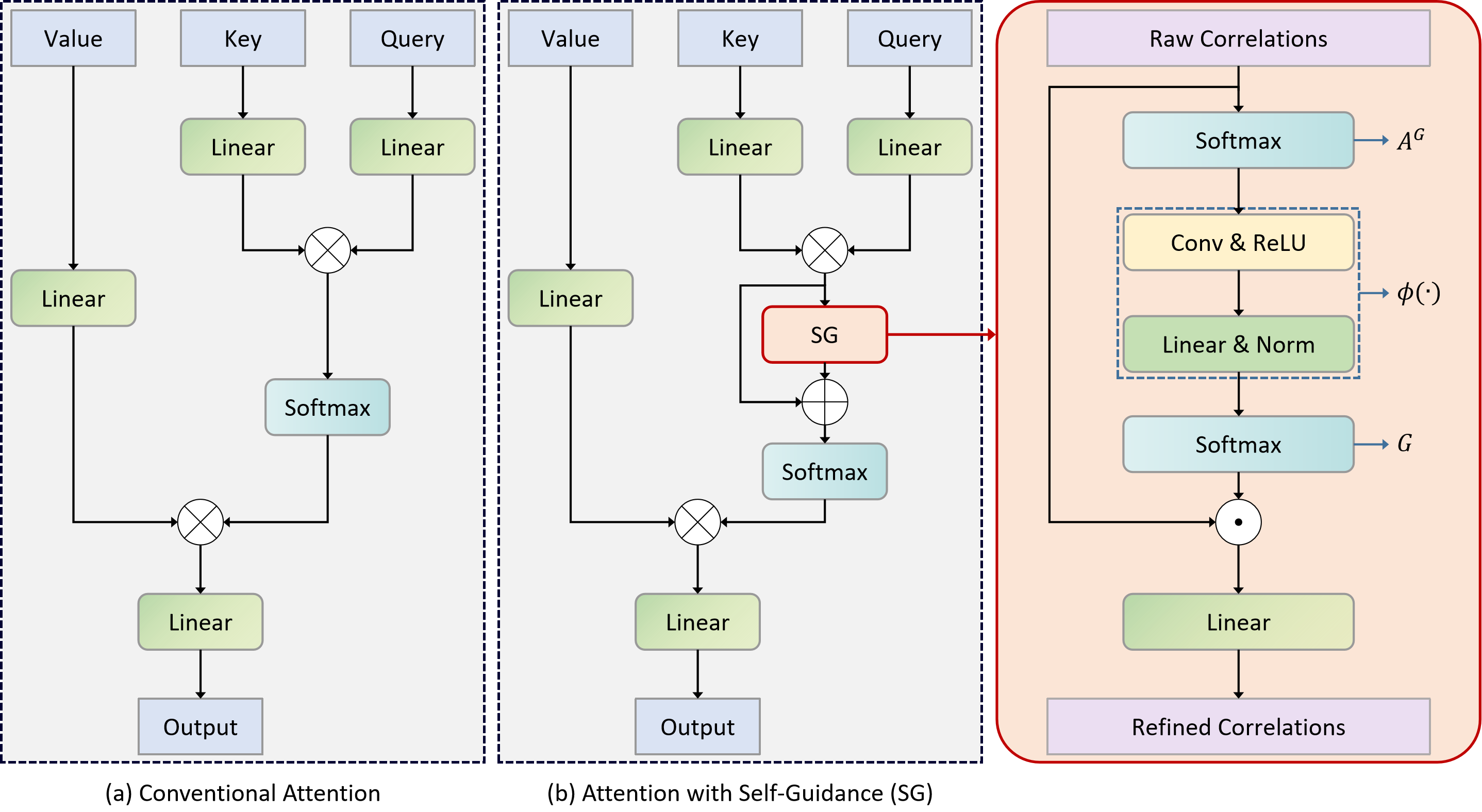}
    \caption{Structures of the conventional attention and the proposed self-guidance module. $\odot$ denotes the element-wise multiplication and $\oplus$ denotes the element-wise addition.}
    \label{fig:5}
\end{figure}

Considering that the image regions of symbols attended by multiple heads should be consistent, we propose to refine raw correlations by eliminating inconsistencies of different attention heads. As shown in Fig. \ref{fig:5}, given raw correlations $\mathbf{E}_1,\ldots,\mathbf{E}_h$ for $h$ heads, the attention guidance can be derived as the concatenation of attention maps computed in Eq. \ref{Softmax}. Afterwards, the guidance map $\mathbf{G}$ is obtained by Eq. \ref{ConvSoftmax}:
\begin{equation}
    \mathbf{G}=\mathrm{softmax}(\phi(\left[\mathbf{A}_1;\ldots;\mathbf{A}_h\right]))
\end{equation}
where $\mathbf{A}_1,\ldots,\mathbf{A}_h$ are attention weights computed from different subspaces, since each $\mathrm{Head}_i$ learns its own set of linear projection matrices $\mathbf{W}_i^Q,\mathbf{W}_i^K$ for queries and keys. Correct activations of symbols should be consistent across multiple heads otherwise they might be noise. $\mathbf{G}$ is obtained by finding inconsistencies through the convolution function $\phi(\cdot)$. The self-guidance module then generates residual correlations to obtain the refined correlations $\mathbf{\hat{E}}$ in Eq. \ref{GuideFunc}:
\begin{equation}\begin{aligned}\label{SelfGuidance}
    \mathbf{\hat{E}}&=\mathbf{E}+\mathrm{SelfGuide}(\mathbf{E},\mathbf{G})\\&=\mathbf{E}+(\mathbf{E}\odot\mathbf{G})\mathbf{W}^G
\end{aligned}\end{equation}
where $\mathbf{W}^G\in\mathbb{R}^{h\times h}$ is a trainable projection matrix for adjusting the guided correlations.

In this way, self-guidance can be readily applied to multi-head attention blocks. Self-guidance facilitates feature aggregation of cross-attention modules since misleading visual features are explicitly suppressed. However, the raw attention is self-refined and lacks external guidance, which can be improved by the neighbor-guidance approach.

\subsubsection{Neighbor-guidance}\label{3-2-2}

\begin{figure}[!t]
    \centering
    \includegraphics[width=0.80\linewidth]{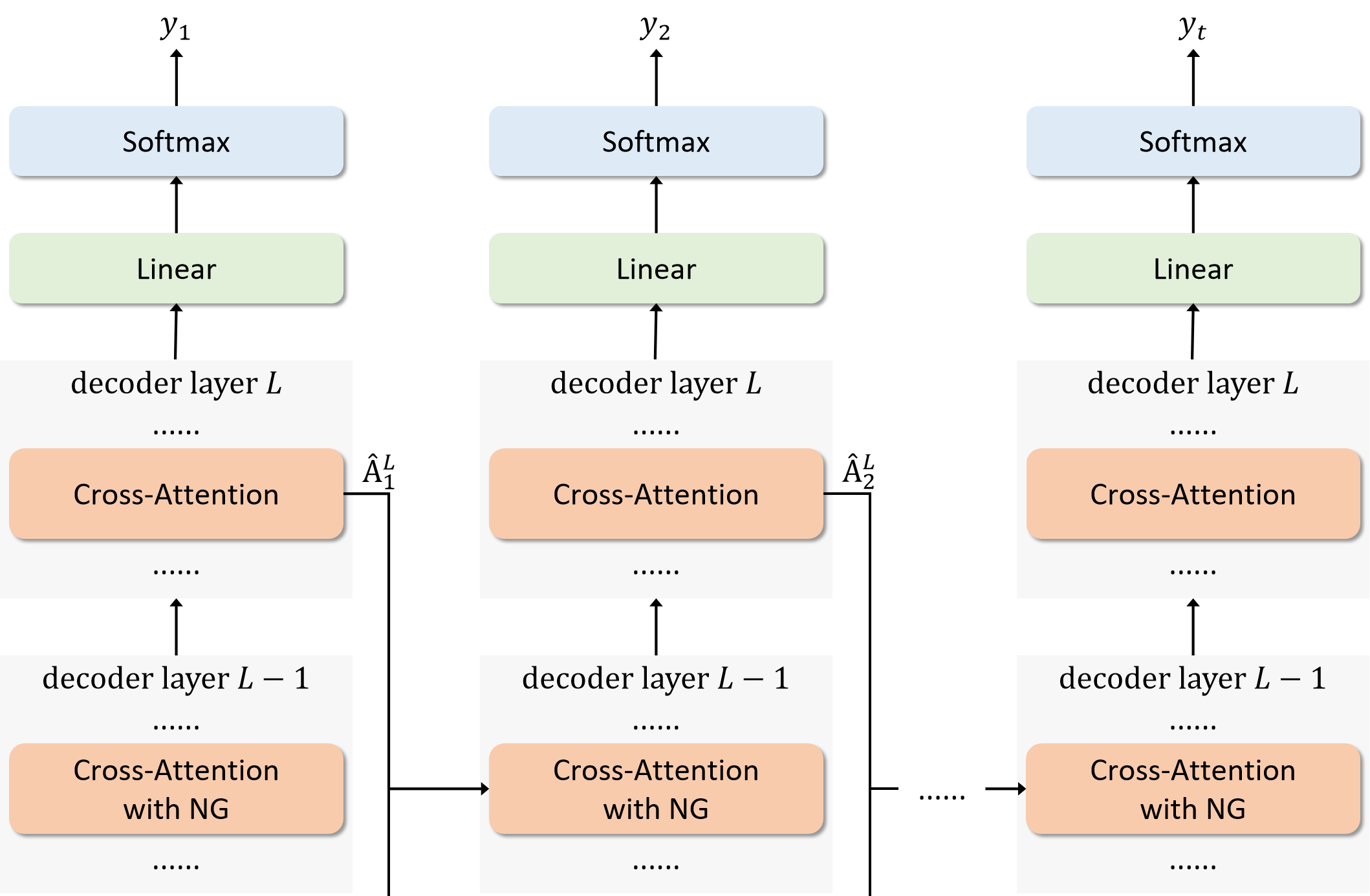}
    \caption{HMER decoder with the proposed neighbor-guidance (NG), which leverages the final attention weights of the previous decoding step to enhance appropriate correlations for middle layers at the current step.}
    \label{fig:6}
\end{figure}

For HMER methods with a stacked decoder \cite{BTTR, CoMER}, each symbol is aligned depending on the alignment of its previously decoded neighbor. Motivated by this, we propose to leverage the final attention weights of the previous decoding step to refine the raw correlations of the current step. This process is illustrated in Fig. \ref{fig:6}. Let $L$ denote the number of decoder layers, given the final attention weights $\mathbf{\hat{A}}^L_{t-1}$ generated in the last layer $L$ at step $t-1$, the guidance map $\mathbf{G}^{l}_t$ for middle layers $l<L$ can be computed as:
\begin{equation}
    \mathbf{G}^{l}_t=\mathrm{mean}(\mathbf{\hat{A}}^L_{t-1})
\end{equation}
where $\mathbf{\hat{A}}^L_{t-1}\in\mathbb{R}^{L\times h}$ is the attention guidance, and $\mathrm{mean}(\cdot)$ represents the average of the $h$ heads, which can be regarded as seeking consensus among multiple attention heads. Similar to the self-guidance module, residual correlations are then generated to obtain the refined correlations $\mathbf{\hat{E}}^{l}_t$:
\begin{equation}\begin{aligned}\label{NeighborGuidance}
    \mathbf{\hat{E}}^{l}_t&=\mathbf{E}^{l}_t+\mathrm{NeighborGuide}(\mathbf{E}^{l}_t,\mathbf{G}^{l}_t)\\&=\mathbf{E}^{l}_t+\alpha(\mathbf{E}^{l}_t\odot\mathbf{G}^{l}_t)
\end{aligned}\end{equation}
where $\alpha$ is a factor for adjusting the guided correlations, which is set to $\left[1,5\right]$ empirically. Note that all the components of neighbor-guidance do not require training, making it convenient to apply to stacked decoders. With the help of neighbor-guidance, the attention weights of the middle layers are refined, thus improving the attention quality of both the current and final layers.

Unlike self-guidance, neighbor-guidance cannot be applied to the last decoder layer. In the middle layers, neighbor-guidance and self-guidance can be executed sequentially to obtain the refined attention weights $\mathbf{\hat{A}}$. If self-guidance is conducted first, let $\mathbf{\hat{E}}^{self}$ denote the correlation map refined by Eq. \ref{SelfGuidance}, and $\mathbf{G}^{neighbor}$ denote the guidance map of neighbor-guidance, then $\mathbf{\hat{A}}$ can be computed as:
\begin{equation}
    \mathbf{\hat{A}}=\mathrm{softmax}(\mathbf{\hat{E}}^{self}+\mathrm{NeighborGuide}(\mathbf{\hat{E}}^{self},\mathbf{G}^{neighbor}))
\end{equation}
while in the final layer, only self-guidance is executed to suppress erroneous correlations. In conclusion, self- and neighbor-guidance are complementary and contribute to feature aggregation and information propagation during the decoding process.

\section{Experiments}\label{sec4}

\subsection{Implementation details}

We employ a DenseNet encoder following the same setting with \cite{BTTR} and \cite{CoMER} for the encoder part. The encoder contains three DenseNet blocks, each consisting of $D=16$ bottleneck layers. The spatial and channel sizes of feature maps are reduced by $\theta=0.5$ through the transition layer between every two DenseNet blocks. The growth rate and dropout rate are set to $k=24$ and $p=0.2$, respectively.

The Transformer decoder consists of $L=3$ decoder layers, where the model dimension is set to $d_{\mathrm{model}}=256$, the dimension size of feed-forward layers is set to $d_{\mathrm{ff}}=1024$, and the dropout rate is set to $0.3$. We use $h=8$ heads in both self- and cross-attention modules. As in CoMER, an attention refinement module is used to model the coverage mechanism. For the proposed attention guidance method, we employ self-guidance modules in the $2^{nd}$ and $3^{rd}$ layers, while neighbor-guidance is only used in the $2^{nd}$ layer. The kernel size $k_c$ and intermediate dimension $d_c$ of the convolution function $\phi(\cdot)$ are set to $k_c=5$ and $d_c=32$, respectively.

The model is trained with the same bidirectional strategy as CoMER. We augment input images using scale augmentation \cite{ScaleAug} with scaling factors uniformly sampled from $\left[0.7,1.4\right]$. The batch size is set to $8$. For the optimizer, we use SGD with a learning rate of 0.08. The weight decay is set to $10^{-4}$, and the momentum is set to $0.9$. The model is implemented with PyTorch 
\cite{PyTorch} framework, and all experiments are conducted on a single NVIDIA GeForce RTX 3090 GPU with 24GB RAM.

During the inference phase, we maintain a list of refined attention weights for neighbor-guidance. New attention matrices are inserted at the end of the list after each decoding step. Besides, we perform an approximate joint search (AJS) \cite{AJS} used in CoMER to leverage the bidirectional information.

\subsection{Datasets and metrics}

To explore the effectiveness of the proposed method, we conduct our experiments on the CROHME datasets \cite{CROHME2014, CROHME2016, CROHME2019}, which are the most widely used public datasets in the HMER task. The training set contains 8836 training samples, while the three test sets CROHME 2014/2016/2019 have 986/1147/1199 test samples. CROHME 2014 test set is used as the validation set to select the best model during the training phase and the adjustment factor $\alpha$ of neighbor-guidance.

For evaluation, we first convert the generated \LaTeX\ sequences into symbol label graph (symLG) format through the evaluation tool officially provided by the CROHME 2019 \cite{CROHME2019} organizers. Evaluation metrics are then reported by the lgEval library \cite{LgEval}, of which expression recognition rate (ExpRate) is the most widely used metric to measure HMER model performance, defined as the percentage of correctly recognized expressions. We use ``ExpRate'' and ``$\leq1$ error'' to represent ExpRates that tolerate 0 to 1 symbol or structural error. Furthermore, the structure recognition rate (StruRate) is used to evaluate the alignment accuracy of the attention model, defined as the percentage of predicted sequences with no structural errors.

\begin{table}[!t]
\centering
\setlength{\abovecaptionskip}{0cm}
\setlength{\belowcaptionskip}{0.2cm}
\caption{Ablation study on the CROHME datasets (in \%). ``Self'' and ``Neighbor'' represent the proposed self-guidance and neighbor-guidance. Neighbor-guidance is applied first when both approaches are used together. The adjustment factor of neighbor-guidance is set to $\alpha=2.5$ for all the cases.}
\label{tab:1}
\resizebox{0.65\linewidth}{!}{
\begin{tabular}{@{}c c c l l@{}}
\toprule
Dataset                    & Self         & Neighbor     & \multicolumn{1}{c}{ExpRate} & \multicolumn{1}{c}{StruRate} \\ \midrule
\multirow{4}{*}{CROHME 14} & -            & -            & 58.93                       & 76.47                        \\
                           & $\checkmark$ & -            & 59.84 (+0.91)               & 77.48 (+1.01)                \\
                           & -            & $\checkmark$ & 59.84 (+0.91)               & \textbf{78.60} (+2.13)       \\
                           & $\checkmark$ & $\checkmark$ & \textbf{60.65} (+1.72)      & 77.99 (+1.52)                \\ \midrule
\multirow{4}{*}{CROHME 16} & -            & -            & 59.46                       & 80.82                        \\
                           & $\checkmark$ & -            & 61.20 (+1.74)               & 81.08 (+0.26)                \\
                           & -            & $\checkmark$ & 59.98 (+0.52)               & 81.34 (+0.52)                \\
                           & $\checkmark$ & $\checkmark$ & \textbf{61.99} (+2.53)      & \textbf{82.04} (+1.22)       \\ \midrule
\multirow{4}{*}{CROHME 19} & -            & -            & 62.89                       & 81.90                        \\
                           & $\checkmark$ & -            & 62.97 (+0.08)               & 81.90 (+0.00)                \\
                           & -            & $\checkmark$ & \textbf{63.55} (+0.66)      & 82.32 (+0.42)                \\
                           & $\checkmark$ & $\checkmark$ & 63.30 (+0.41)               & \textbf{82.40} (+0.50)       \\ \bottomrule
\end{tabular}}
\end{table}

\subsection{Ablation study}

We perform ablation experiments on the CROHME datasets to verify the effectiveness of our method, as shown in Table \ref{tab:1}. We re-implement the CoMER model as our baseline and achieve results similar to those originally reported in \cite{CoMER}. The proposed method significantly improves the performance of the baseline model. Comparing the results in Table \ref{tab:1}, we can observe that:

(1) When self-guidance is used to refine the attention maps, the ExpRates are improved on all three datasets, while the StruRates are improved slightly. This suggests that self-guidance effectively improves attention quality, but its improvement in aligning expression structures is limited due to the lack of external guidance.

(2) Compared to self-guidance, neighbor-guidance can bring higher StruRate improvement but less ExpRate gains. This suggests that using the alignment information from the previous step as attention guidance can effectively improve the alignment results. However, it is also limited since it can only be applied to the middle layers.

(3) The fusion of the two approaches achieves consistent improvements in both ExpRate and StruRate. Especially it achieves an average ExpRate improvement of 1.55\% compared to the baseline model, which is much better than using only one approach. This experiment demonstrates that self-guidance and neighbor-guidance are complementary and can work together effectively to improve attention results.

\begin{table}[!t]
\centering
\setlength{\abovecaptionskip}{0cm}
\setlength{\belowcaptionskip}{0.2cm}
\caption{Comparison of ExpRate (\%) between different settings of neighbor-guidance.}
\label{tab:2}
\resizebox{0.70\linewidth}{!}{
\begin{tabular}{@{}c c c c c@{}}
\toprule
Guidance order & $\alpha$ & CROHME 14      & CROHME 16      & CROHME 19      \\ \midrule
Neighbor first & 2.5      & 60.65          & \textbf{61.99} & \textbf{63.30} \\
Self first     & 1        & 60.65          & 61.73          & 63.05          \\
Self first     & 2.5      & 60.75          & 61.81          & \textbf{63.30} \\
Self first     & 5        & \textbf{60.85} & 61.90          & 62.64          \\ \bottomrule
\end{tabular}}
\end{table}

\begin{table}[!t]
\centering
\setlength{\abovecaptionskip}{0cm}
\setlength{\belowcaptionskip}{0.2cm}
\caption{Performance comparison (in \%) with previous state-of-the-art methods on the CROHME datasets. $\ast$ denotes results using scale augmentation \cite{ScaleAug}, and $\dag$ denotes the original reported results of BTTR \cite{BTTR} and CoMER \cite{CoMER}. ``AttnGuide'' indicates our proposed method based on the CoMER model.}
\label{tab:3}
\resizebox{0.88\linewidth}{!}{
\begin{tabular}{@{}c c c c c c c@{}}
\toprule
\multirow{2}{*}{Methods} & \multicolumn{2}{c}{CROHME 2014} & \multicolumn{2}{c}{CROHME 2016} & \multicolumn{2}{c}{CROHME 2019} \\ \cmidrule(l){2-7} 
                         & ExpRate        & $\leq1$ error  & ExpRate        & $\leq1$ error  & ExpRate        & $\leq1$ error  \\ \midrule
UPV \cite{CROHME2014}    & 37.22          & 44.22          & -              & -              & -              & -              \\
TOKYO \cite{CROHME2016}  & -              & -              & 43.94          & 50.91          & -              & -              \\
PAL \cite{PAL2018}       & 39.66          & 56.80          & -              & -              & -              & -              \\
WAP \cite{WAP}           & 46.55          & 61.16          & 44.55          & 57.10          & -              & -              \\
PAL-v2 \cite{PAL}        & 48.88          & 64.50          & 49.61          & 64.08          & -              & -              \\
DLA \cite{DLA}           & 49.85          & -              & 47.34          & -              & -              & -              \\
DWAP \cite{DWAP}         & 50.10          & -              & 47.50          & -              & -              & -              \\
DWAP-TD \cite{TD}        & 49.10          & 64.20          & 48.50          & 62.30          & 51.40          & 66.10          \\
WS-WAP \cite{WS-WAP}     & 53.65          & -              & 51.96          & 64.34          & -              & -              \\
TDv2 \cite{TDv2}         & 53.62          & -              & 55.18          & -              & 58.72          & -              \\
ABM \cite{ABM}           & 56.85          & 73.73          & 52.92          & 69.66          & 53.96          & 71.06          \\
CAN-DWAP \cite{CAN}      & 57.00          & 74.21          & 56.06          & 71.49          & 54.88          & 71.98          \\
CAN-ABM \cite{CAN}       & 57.26          & 74.52          & 56.15          & 72.71          & 55.96          & 72.73          \\
SAN \cite{SAN}           & 56.20          & 72.60          & 53.60          & 69.60          & 53.50          & 69.30          \\
GETD \cite{GETD}         & 53.45          & 67.54          & 55.27          & 68.43          & 54.13          & 67.72          \\
Li et al.$^\ast$ \cite{ScaleAug} & 56.59  & 69.07          & 54.58          & 69.31          & -              & -              \\
Ding et al.$^\ast$ \cite{StackedDecoder}  & 58.72   & -    & 57.72          & 70.01          & 61.38          & 75.15          \\
BTTR$^\dag$ \cite{BTTR}  & 53.96          & 66.02          & 52.31          & 63.90          & 52.96          & 65.97          \\
BTTR$^\ast$ \cite{BTTR}  & 55.17          & 67.85          & 56.58          & 68.88          & 59.55          & 72.23          \\
CoMER$^{\ast\dag}$ \cite{CoMER} & 59.33   & 71.70          & 59.81          & 74.37          & 62.97          & 77.40          \\ \midrule
CoMER \cite{CoMER}       & 55.17          & 67.24          & 56.15          & 70.62          & 59.38          & 73.81          \\
AttnGuide                & \textbf{57.61} & 68.86          & \textbf{58.24} & 70.71          & \textbf{60.47} & 74.48          \\ \midrule
CoMER$^{\ast}$ \cite{CoMER} & 58.93       & 70.49          & 59.46          & 73.58          & 62.89          & 77.65          \\
AttnGuide$^\ast$         & \textbf{60.65} & 72.01          & \textbf{61.99} & 75.07          & \textbf{63.30} & 77.73          \\ \bottomrule
\end{tabular}}
\end{table}

Besides, we also explore the impact of different settings of neighbor-guidance, as shown in Table \ref{tab:2}. As we can see, the results obtained by the fusion methods are almost the same in terms of the order of neighbor-guidance and self-guidance. Comparing the last three rows, it should be noted that with the increase of $\alpha$, the model cannot always obtain performance improvements. When $\alpha=5$, the ExpRate on CROHME 2019 is even lower than that of $\alpha=2.5$. This is caused by redundant information in guidance maps since the final attention weights from the previous step are not necessarily accurate. Empirically, $\alpha$ is set to $2.5$ to achieve the best performance.

\subsection{Comparison with state-of-the-art methods}

We compare the proposed method with existing state-of-the-art methods, as shown in Table \ref{tab:3}. The original results of CoMER are obtained using scale augmentation \cite{ScaleAug}, while most previous methods do not use data augmentation. Therefore, we compare our method using scale augmentation with other previous methods using scale augmentation. To fairly compare with previous methods without data augmentation, we re-train our method and the baseline CoMER model without using scale augmentation and set the beam size of AJS \cite{AJS} to 1 during the inference phase.

As shown in Table \ref{tab:3}, with adopting CoMER as the baseline, our method achieves state-of-the-art ExpRate performance on the CROHME datasets, whether or not data augmentation is used. We improve the baseline SOTA model CoMER on CROHME 2016 by a significant margin of 2.18\% over its original reported result.

\subsection{Case study with maps}

\begin{figure}[!t]
    \centering
    \includegraphics[width=0.80\linewidth]{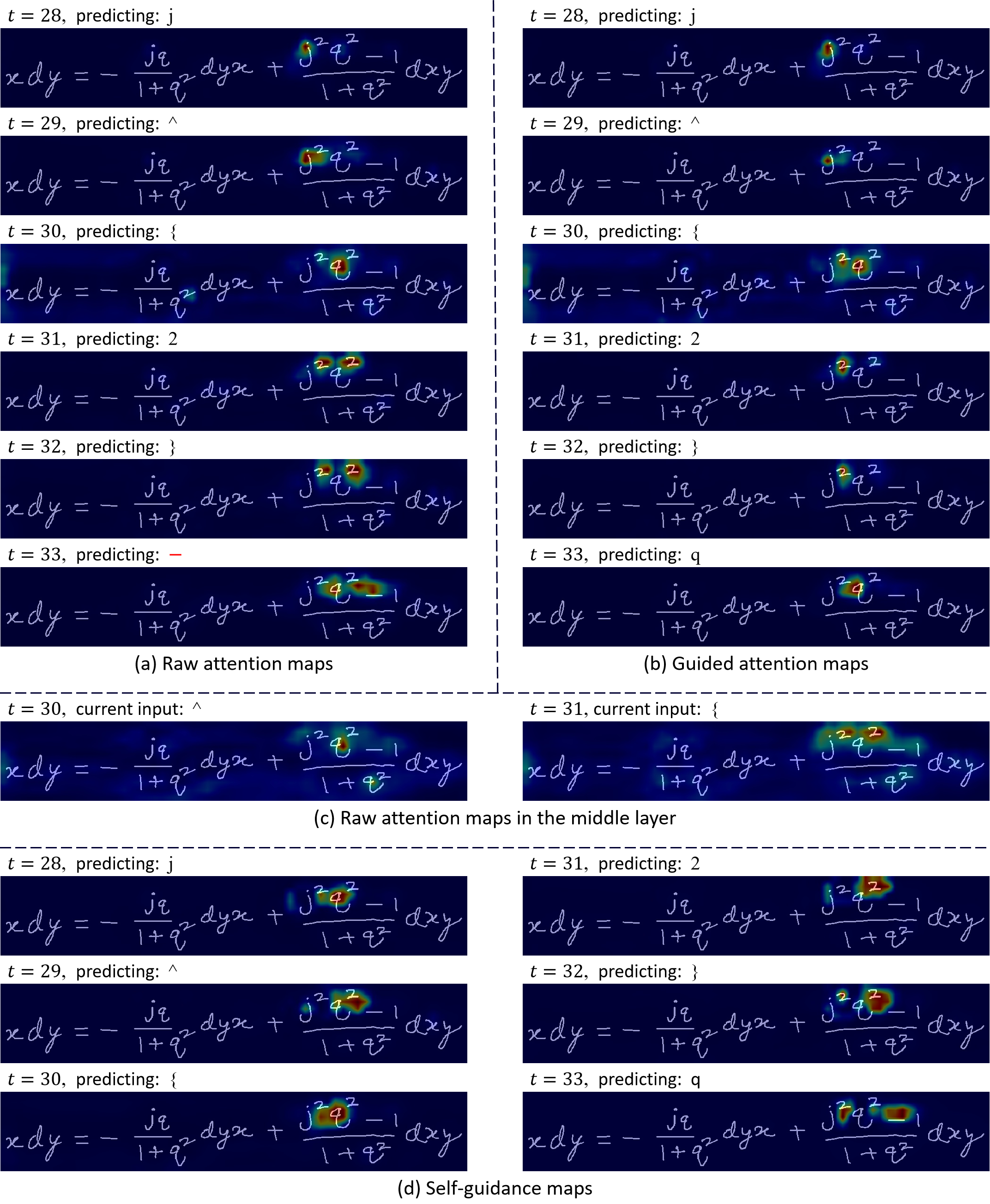}
    \caption{Visualization of attention maps corrected by attention guidance. Note that the refined attention map at each step $t$ is reused as the neighbor-guidance map for step $t+1$.}
    \label{fig:7}
\end{figure}

This section shows some qualitative recognition results of the proposed attention guidance method. Fig. \ref{fig:7} illustrates the attention maps before and after attention guidance. As shown in subfig (a), the attention module without guidance fails to distinguish between the two superscripts of ``$j^{2}$'' and ``$q^{2}$''. The attention incorrectly focuses on the image areas around ``$q^{2}$'' after step $t=30$. As shown in subfig (c), this is caused by the context leakage phenomenon where the middle layer mistakenly activates the areas around ``$q^{2}$''. As a result, the information behind the intended context is incorporated in the attention result, thus confusing the generation process after step $t=30$. This issue is solved with the help of self-guidance and neighbor-guidance, as shown in subfig (b). Compared with subfig (a), erroneous activations on ``$q^2$'' are suppressed at step $t=31$ and $t=32$. The model can make the correct prediction based on the input context even if the erroneous activations are not entirely removed at step $t=31$. In subfig (b), the attention map at step $t$ is also the neighbor-guidance map for step $t+1$. Subfig (d) shows the guidance maps derived from the self-guidance module. As we can see, the image regions of incorrect activations in the original attention maps are assigned high weights. Using these guidance maps, erroneous correlations can be removed through the adjustment operation in Eq. \ref{SelfGuidance}. In conclusion, the visualization experiment shows that our proposed attention guidance method can effectively alleviate the context leakage phenomenon.

\begin{figure}[!t]
    \centering
    \includegraphics[width=0.95\linewidth]{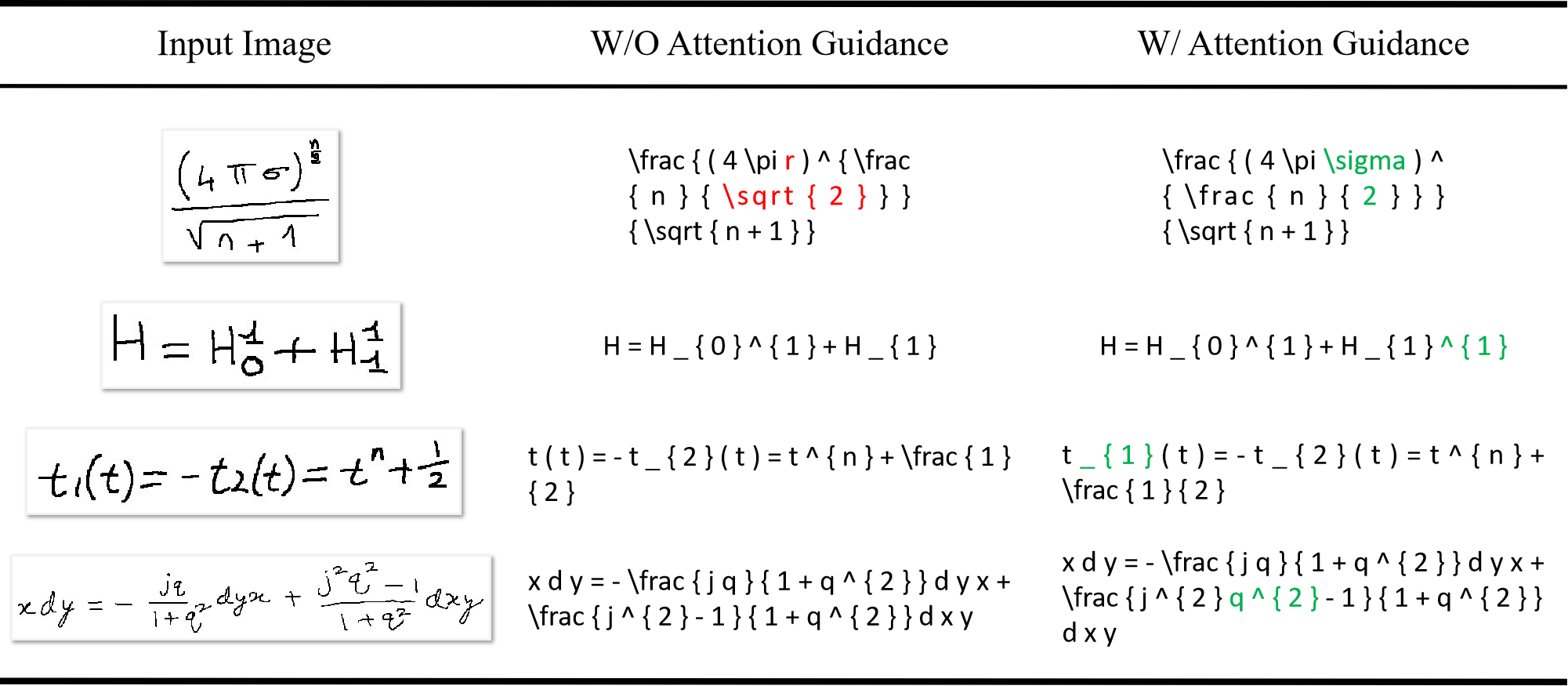}
    \caption{Recognition errors corrected by attention guidance. The red symbols are recognition errors of our baseline model CoMER \cite{CoMER}, and the green symbols are the recognition results after attention guidance.}
    \label{fig:8}
\end{figure}

Fig. \ref{fig:8} further illustrates some examples in the CROHME test sets. In the first example, the incorrectly inserted ``\textbackslash sqrt \{ $\cdot$ \}'' is due to the misalignment on image regions of ``$\sqrt{n+1}$'', which should be parsed in the future step. Similar context leakage phenomenons arise in the last three examples, causing symbols in the predicted sequences to be omitted. It can be seen that our proposed attention guidance mechanism successfully corrects the recognition errors brought by the context leakage phenomenon.

\section{Conclusion}\label{sec5}

Misalignment, especially under-parsing, remains a challenging problem in HMER. We observe that one reason is the context leakage phenomenon. To address this issue, we propose an attention guidance mechanism that aims to suppress erroneous attention explicitly. In this way, information after the intended context is suppressed in the attention result. This paper has four main contributions: (1) We propose the context leakage phenomenon and a general attention guidance mechanism to solve this issue. (2) We devise self-guidance to refine the raw attention by seeking consensus among multiple attention heads since the image regions of symbols attended by multiple heads should be consistent. (3) Depending on the observation of the alignment process of the stacked decoder, we devise neighbor-guidance to reuse attention weights from the previous decoding step to refine the raw attention in the current step. (4) Experiments on standard benchmarks show that our method outperforms state-of-the-art methods, achieving ExpRates of 60.75\% / 61.81\% / 63.30\% on the CROHME 2014/ 2016/ 2019 datasets.

The proposed attention guidance mechanism not only works for the HMER task. The ideas of self-guidance and neighbor-guidance can help improve grounding quality and alignment results for other tasks requiring dynamic alignment. In future work, we intend to apply our attention guidance mechanism to solve tasks such as document understanding, image captioning, and machine translation. Besides, we also intend to explore other attention guidance approaches to facilitate information propagation in attention-based models further.

\bibliographystyle{elsarticle-num} 
\bibliography{bibliography}

\begin{thebibliography}{10}
\expandafter\ifx\csname url\endcsname\relax
  \def\url#1{\texttt{#1}}\fi
\expandafter\ifx\csname urlprefix\endcsname\relax\def\urlprefix{URL }\fi
\expandafter\ifx\csname href\endcsname\relax
  \def\href#1#2{#2} \def\path#1{#1}\fi

\bibitem{IJDAR2012a}
R.~Zanibbi, D.~Blostein, Recognition and retrieval of mathematical expressions, International Journal on Document Analysis and Recognition 15~(4) (2012) 331--357.

\bibitem{WAP}
J.~Zhang, J.~Du, S.~Zhang, D.~Liu, Y.~Hu, J.-S. Hu, S.~Wei, L.-R. Dai, Watch, attend and parse: An end-to-end neural network based approach to handwritten mathematical expression recognition, Pattern Recognition 71 (2017) 196--206.

\bibitem{DWAP}
J.~Zhang, J.~Du, L.~Dai, Multi-scale attention with dense encoder for handwritten mathematical expression recognition, in: 24th International Conference on Pattern Recognition, 2018, pp. 2245--2250.

\bibitem{TD}
J.~Zhang, J.~Du, Y.~Yang, Y.~Song, S.~Wei, L.~Dai, A tree-structured decoder for image-to-markup generation, in: Proceedings of the 37th International Conference on Machine Learning, Vol. 119, 2020, pp. 11076--11085.

\bibitem{BTTR}
W.~Zhao, L.~Gao, Z.~Yan, S.~Peng, L.~Du, Z.~Zhang, Handwritten mathematical expression recognition with bidirectionally trained transformer, in: 16th International Conference on Document Analysis and Recognition, Vol. 12822, 2021, pp. 570--584.

\bibitem{CoMER}
W.~Zhao, L.~Gao, {CoMER}: Modeling coverage for transformer-based handwritten mathematical expression recognition, in: Proceedings of the European Conference on Computer Vision, Vol. 13688, 2022, pp. 392--408.

\bibitem{TDv2}
C.~Wu, J.~Du, Y.~Li, J.~Zhang, C.~Yang, B.~Ren, Y.~Hu, {TD}v2: A novel tree-structured decoder for offline mathematical expression recognition, in: Thirty-Sixth {AAAI} Conference on Artificial Intelligence, 2022, pp. 2694--2702.

\bibitem{ABM}
X.~Bian, B.~Qin, X.~Xin, J.~Li, X.~Su, Y.~Wang, Handwritten mathematical expression recognition via attention aggregation based bi-directional mutual learning, in: Thirty-Sixth {AAAI} Conference on Artificial Intelligence, 2022, pp. 113--121.

\bibitem{CAN}
B.~Li, Y.~Yuan, D.~Liang, X.~Liu, Z.~Ji, J.~Bai, W.~Liu, X.~Bai, When counting meets hmer: Counting-aware network for handwritten mathematical expression recognition, in: Proceedings of the European Conference on Computer Vision, Vol. 13688, 2022, pp. 197--214.

\bibitem{GCN}
X.~Zhang, H.~Ying, Y.~Tao, Y.~Xing, G.~Feng, General category network: Handwritten mathematical expression recognition with coarse-grained recognition task, in: International Conference on Acoustics, Speech and Signal Processing, 2023, pp. 1--5.

\bibitem{EMNLP23}
Z.~Chen, J.~Han, C.~Yang, Y.~Zhou, Language model is suitable for correction of handwritten mathematical expressions recognition, in: Proceedings of the 2023 Conference on Empirical Methods in Natural Language Processing, 2023, pp. 4057--4068.

\bibitem{TMM23}
L.~Zhe, W.~Xinyu, L.~Yuliang, J.~Lianwen, H.~Yichao, D.~Kai, Improving handwritten mathematical expression recognition via similar symbol distinguishing, {IEEE} Transactions on Multimedia 26 (2023) 90--102.

\bibitem{ABINetPP}
S.~Fang, Z.~Mao, H.~Xie, Y.~Wang, C.~Yan, Y.~Zhang, Abinet++: Autonomous, bidirectional and iterative language modeling for scene text spotting, {IEEE} Transactions on Pattern Analysis and Machine Intelligence 45 (2023) 7123--7141.

\bibitem{LISTER}
C.~Cheng, P.~Wang, C.~Da, Q.~Zheng, C.~Yao, {LISTER}: Neighbor decoding for length-insensitive scene text recognition, CoRR abs/2308.12774 (2023).

\bibitem{StackedDecoder}
H.~Ding, K.~Chen, Q.~Huo, An encoder-decoder approach to handwritten mathematical expression recognition with multi-head attention and stacked decoder, in: 16th International Conference on Document Analysis and Recognition, 2021, pp. 602--616.

\bibitem{AoA}
L.~Huang, W.~Wang, J.~Chen, X.~Wei, Attention on attention for image captioning, in: International Conference on Computer Vision, 2019, pp. 4633--4642.

\bibitem{ProphetAttn}
F.~Liu, X.~Ren, X.~Wu, S.~Ge, W.~Fan, Y.~Zou, X.~Sun, Prophet attention: Predicting attention with future attention, in: Advances in Neural Information Processing Systems, Vol.~33, 2020, pp. 1865--1876.

\bibitem{CROHME2014}
H.~Mouchere, C.~Viard-Gaudin, R.~Zanibbi, U.~Garain, {ICFHR} 2014 competition on recognition of on-line handwritten mathematical expressions ({CROHME} 2014), in: 14th International Conference on Frontiers in Handwriting Recognition, 2014, pp. 791--796.

\bibitem{CROHME2016}
H.~Mouchere, C.~Viard-Gaudin, R.~Zanibbi, U.~Garain, {ICFHR}2016 {CROHME}: Competition on recognition of online handwritten mathematical expressions, in: 15th International Conference on Frontiers in Handwriting Recognition, 2016, pp. 607--612.

\bibitem{CROHME2019}
M.~Mahdavi, R.~Zanibbi, H.~Mouch{\`{e}}re, C.~Viard{-}Gaudin, U.~Garain, {ICDAR} 2019 {CROHME} + {TFD:} competition on recognition of handwritten mathematical expressions and typeset formula detection, in: 2019 International Conference on Document Analysis and Recognition, 2019, pp. 1533--1538.

\bibitem{IJDAR2000}
K.-F. Chan, D.-Y. Yeung, Mathematical expression recognition: A survey, International Journal on Document Analysis and Recognition (2000) 3–15.

\bibitem{PAMI2002}
R.~Zanibbi, D.~Blostein, J.~Cordy, Recognizing mathematical expressions using tree transformation, {IEEE} Transactions on Pattern Analysis and Machine Intelligence 24 (2002) 1455–1467.

\bibitem{PRL2014a}
F.~Álvaro, J.-A. Sánchez, J.-M. Benedí, Recognition of on-line handwritten mathematical expressions using 2d stochastic context-free grammars and hidden markov models, Pattern Recognition Letters (2014) 58–67.

\bibitem{PRL2014b}
A.-M. Awal, H.~Mouchère, C.~Viard-Gaudin, A global learning approach for an online handwritten mathematical expression recognition system, Pattern Recognition Letters (2014) 68–77.

\bibitem{PR2016}
F.~Álvaro, J.-A. Sánchez, J.-M. Benedí, An integrated grammar-based approach for mathematical expression recognition, Pattern Recognition 51 (2016) 135–147.

\bibitem{IJDAR2020}
F.~D. Julca{-}Aguilar, H.~Mouch{\`{e}}re, C.~Viard{-}Gaudin, N.~S.~T. Hirata, A general framework for the recognition of online handwritten graphics, International Journal on Document Analysis and Recognition 23~(2) (2020) 143--160.

\bibitem{IJDAR2012b}
S.~MacLean, G.~Labahn, A new approach for recognizing handwritten mathematics using relational grammars and fuzzy sets, International Journal on Document Analysis and Recognition 16~(2) (2013) 139--163.

\bibitem{ShowAttendTell}
K.~Xu, J.~Ba, R.~Kiros, K.~Cho, A.~C. Courville, R.~Salakhutdinov, R.~S. Zemel, Y.~Bengio, Show, attend and tell: Neural image caption generation with visual attention, in: Proceedings of the 32nd International Conference on Machine Learning, Vol.~37, 2015, pp. 2048--2057.

\bibitem{FAN-STR}
Z.~Cheng, F.~Bai, Y.~Xu, G.~Zheng, S.~Pu, S.~Zhou, Focusing attention: Towards accurate text recognition in natural images, in: {IEEE} International Conference on Computer Vision, 2017, pp. 5086--5094.

\bibitem{ONN-HTR}
H.~H. Mohammed, J.~Malik, S.~Al{-}M{\'{a}}adeed, S.~Kiranyaz, 2d self-organized {ONN} model for handwritten text recognition, Applied Soft Computing 127 (2022) 109311.

\bibitem{Coarse2Fine}
Y.~Deng, A.~Kanervisto, J.~Ling, A.~M. Rush, Image-to-markup generation with coarse-to-fine attention, in: Proceedings of the 34th International Conference on Machine Learning, Vol.~70, 2017, pp. 980--989.

\bibitem{GRU}
J.~Chung, {\c{C}}.~G{\"{u}}l{\c{c}}ehre, K.~Cho, Y.~Bengio, Empirical evaluation of gated recurrent neural networks on sequence modeling, CoRR abs/1412.3555 (2014).

\bibitem{DenseNet}
G.~Huang, Z.~Liu, L.~van~der Maaten, K.~Q. Weinberger, Densely connected convolutional networks, in: 2017 {IEEE} Conference on Computer Vision and Pattern Recognition, 2017, pp. 2261--2269.

\bibitem{VGG}
K.~Simonyan, A.~Zisserman, Very deep convolutional networks for large-scale image recognition, in: 3rd International Conference on Learning Representations, 2015.

\bibitem{Transformer}
A.~Vaswani, N.~Shazeer, N.~Parmar, J.~Uszkoreit, L.~Jones, A.~N. Gomez, L.~u. Kaiser, I.~Polosukhin, Attention is all you need, in: Advances in Neural Information Processing Systems, Vol.~30, 2017.

\bibitem{FasterRCNN}
S.~Ren, K.~He, R.~B. Girshick, J.~Sun, Faster {R-CNN:} towards real-time object detection with region proposal networks, {IEEE} Transactions on Pattern Analysis and Machine Intelligence 39~(6) (2017) 1137--1149.

\bibitem{ScaleAug}
Z.~Li, L.~Jin, S.~Lai, Y.~Zhu, Improving attention-based handwritten mathematical expression recognition with scale augmentation and drop attention, in: 17th International Conference on Frontiers in Handwriting Recognition, 2020, pp. 175--180.

\bibitem{PAL}
J.~Wu, F.~Yin, Y.~Zhang, X.~Zhang, C.~Liu, Handwritten mathematical expression recognition via paired adversarial learning, International Journal of Computer Vision 128~(10) (2020) 2386--2401.

\bibitem{WS-WAP}
T.~Truong, C.~T. Nguyen, K.~M. Phan, M.~Nakagawa, Improvement of end-to-end offline handwritten mathematical expression recognition by weakly supervised learning, in: 17th International Conference on Frontiers in Handwriting Recognition, 2020, pp. 181--186.

\bibitem{ReLU}
X.~Glorot, A.~Bordes, Y.~Bengio, Deep sparse rectifier neural networks, in: Proceedings of the Fourteenth International Conference on Artificial Intelligence and Statistics, 2011, pp. 315--323.

\bibitem{BatchNorm}
S.~Ioffe, C.~Szegedy, Batch normalization: Accelerating deep network training by reducing internal covariate shift, in: International Conference on Machine Learning, 2015, pp. 448--456.

\bibitem{DETR}
N.~Carion, F.~Massa, G.~Synnaeve, N.~Usunier, A.~Kirillov, S.~Zagoruyko, End-to-end object detection with transformers, in: European Conference on Computer Vision, 2020, pp. 213--229.

\bibitem{VisualGPT}
J.~Chen, H.~Guo, K.~Yi, B.~Li, M.~Elhoseiny, Visualgpt: Data-efficient adaptation of pretrained language models for image captioning, in: {IEEE/CVF} Conference on Computer Vision and Pattern Recognition, 2022, pp. 18009--18019.

\bibitem{TROCR}
M.~Li, T.~Lv, J.~Chen, L.~Cui, Y.~Lu, D.~A.~F. Flor{\^{e}}ncio, C.~Zhang, Z.~Li, F.~Wei, Trocr: Transformer-based optical character recognition with pre-trained models, in: Thirty-Seventh {AAAI} Conference on Artificial Intelligence, 2023, pp. 13094--13102.

\bibitem{PyTorch}
A.~Paszke, S.~Gross, F.~Massa, A.~Lerer, J.~Bradbury, G.~Chanan, T.~Killeen, Z.~Lin, N.~Gimelshein, L.~Antiga, A.~Desmaison, A.~K{\"{o}}pf, E.~Z. Yang, Z.~DeVito, M.~Raison, A.~Tejani, S.~Chilamkurthy, B.~Steiner, L.~Fang, J.~Bai, S.~Chintala, Pytorch: An imperative style, high-performance deep learning library, in: Advances in Neural Information Processing Systems, 2019, pp. 8024--8035.

\bibitem{AJS}
L.~Liu, M.~Utiyama, A.~M. Finch, E.~Sumita, Agreement on target-bidirectional neural machine translation, in: The 2016 Conference of the North American Chapter of the Association for Computational Linguistics, 2016, pp. 411--416.

\bibitem{LgEval}
R.~Zanibbi, H.~Mouch{\`{e}}re, C.~Viard{-}Gaudin, Evaluating structural pattern recognition for handwritten math via primitive label graphs, in: Document Recognition and Retrieval, Vol. 8658, 2013, p. 865817.

\bibitem{PAL2018}
J.~Wu, F.~Yin, Y.~Zhang, X.~Zhang, C.~Liu, Image-to-markup generation via paired adversarial learning, in: Machine Learning and Knowledge Discovery in Databases, Vol. 11051, 2018, pp. 18--34.

\bibitem{DLA}
A.~D. Le, Recognizing handwritten mathematical expressions via paired dual loss attention network and printed mathematical expressions, in: 2020 {IEEE/CVF} Conference on Computer Vision and Pattern Recognition, {CVPR} Workshops 2020, 2020, pp. 2413--2418.

\bibitem{SAN}
Y.~Yuan, X.~Liu, W.~Dikubab, H.~Liu, Z.~Ji, Z.~Wu, X.~Bai, Syntax-aware network for handwritten mathematical expression recognition, in: {IEEE/CVF} Conference on Computer Vision and Pattern Recognition,, 2022, pp. 4543--4552.

\bibitem{GETD}
J.~Tang, H.~Guo, J.~Wu, F.~Yin, L.~Huang, Offline handwritten mathematical expression recognition with graph encoder and transformer decoder, Pattern Recognition 148 (2024) 110155.

\end{thebibliography}

\end{document}